\documentclass[letterpaper]{article} 
\usepackage[submission]{aaai25}  
\usepackage{times}  
\usepackage{helvet}  
\usepackage{courier}  
\usepackage[hyphens]{url}  
\usepackage{graphicx} 
\usepackage{natbib}  
\usepackage{caption} 
\usepackage{algorithm}
\usepackage{algorithmic}
\usepackage{amsmath}
\usepackage{booktabs}       
\usepackage{amsfonts}       
\usepackage{nicefrac}       
\usepackage{microtype}      
\usepackage{xcolor}         
\usepackage{graphicx}
\usepackage{amsthm}
\usepackage{amssymb}
\usepackage{amsmath}
\usepackage{adjustbox}
\usepackage{array}
\usepackage{tabularx}

\usepackage{newfloat}
\usepackage{listings}
\DeclareCaptionStyle{ruled}{labelfont=normalfont,labelsep=colon,strut=off} 
\floatstyle{ruled}
\newfloat{listing}{tb}{lst}{}
\floatname{listing}{Listing}
\title{Disentangled Structural and Featural Representation for Task-Agnostic Graph Valuation}
\author{
    Ali Falahati\textsuperscript{\rm 1}, Mohammad Mohammadi Amiri\textsuperscript{\rm 2}
}
\affiliations{
    \textsuperscript{\rm 1} Max Planck Institute for Software Systems\\
        \textsuperscript{\rm 2} Rensselaer Polytechnic Institute\\

    afalahat@mpi-sws.org, mamiri@rpi.edu
}

\usepackage{bibentry}

\begin{document}

\maketitle

\begin{abstract}
 With the emergence of data marketplaces, the demand for methods to assess the value of data has increased significantly. While numerous techniques have been proposed for this purpose, none have specifically addressed graphs as the main data modality. Graphs are widely used across various fields, ranging from chemical molecules to social networks. In this study, we break down graphs into two main components: structural and featural,  and we focus on evaluating data without relying on specific task-related metrics, making it applicable in practical scenarios where validation requirements may be lacking. We introduce a novel framework called blind message passing, which aligns seller's and buyer's graphs using a shared node permutation based on graph matching. This allows us to utilize the graph Wasserstein distance to quantify the differences in the structural distribution of graph datasets, called the \textit{structural disparities}. We then consider featural aspects of buyer's and seller's graphs for data valuation and capture their statistical similarities and differences, referred to as \textit{relevance} and \textit{diversity}, respectively. Our approach ensures that buyers and sellers remain unaware of each other's datasets. Our experiments on real datasets demonstrate the effectiveness of our approach in capturing the relevance, diversity, and structural disparities of seller data for buyers, particularly in graph-based data valuation scenarios.
\end{abstract}

\section{Introduction}

With the advent of foundation models \cite{foundation}, the demand for large and diverse datasets has increased significantly. Data marketplaces \cite{datamarketplace,marketplace1} have emerged as transformative platforms for exchanging data. These marketplaces allow data owners to sell their datasets and enable buyers to acquire essential data for their analytical and research needs. A notable development in this domain is the emergence of generative models as potential sellers in data marketplaces. Generative models, like generative adversarial networks \cite{gan} and variational autoencoders (Kingma et al. 2013), can create synthetic datasets that mimic real-world data. These models generate high-quality, anonymized data that retains the statistical properties of the original datasets, making them particularly useful in situations where data privacy and scarcity are significant concerns. A critical aspect of these marketplaces is the valuation of data, which determines the quality and desirability of datasets. A data marketplace primarily consists of three entities: data sellers, a broker, and data buyers. Data sellers possess the data and provide it to the broker in exchange for compensation. Data buyers seek to obtain this data, with the broker facilitating the transactions. Given the value of data as a resource, it is crucial to develop a systematic approach to assess the value of the data for both sellers and buyers. This process, known as data valuation, is fundamental to ensuring a fair marketplace for all parties involved. Data valuation is a complex process that involves assessing the quality, relevance, and potential utility of data for specific applications concerning the buyer's demands. It is particularly valuable in fields such as finance, healthcare, marketing, and scientific research, where data-driven insights are crucial.

Data valuation can be performed based on either ``intrinsic'' or ``extrinsic'' factors. Intrinsic data valuation is data-driven and focuses on the quality of the dataset itself \cite{UnlockingValuePrivacyTrading,DataSupportAIAllPricingValuationGovernance}. In contrast, extrinsic data valuation takes into account demand-supply dynamics and game-theoretic mechanisms \cite{DataCollectionWirelessCommunicationIoT,SurveyDataPricingMethods}.
Intrinsic data valuation is often paired with a utility metric for validation \cite{DataShapleyEquitableValuation,EfficientValuationBasedShapley}, or with a specific machine learning (ML) task \cite{datamarketplace,ModelBasedPricingMarketplace}. 
In particular, for ML applications, data valuation frequently relies on the presence of a validation set, with validation accuracy serving as the metric \cite{PrincipledApproachDataValuationForFL,YouLoveTheCore}.
Additionally, the value of training data is often estimated by evaluating ML models trained on a specific target task \cite{SurveyDataPricingEconomicsDataScience,DEALEREnd2EndModelMarketplaceDP}. 
In contrast, extrinsic data valuation techniques take into account external factors like competition and market demand \cite{TowardsDataAuctionsExternalities,InformationSaleCompetition}. This approach involves assessing customer demand for products and analyzing competitors' pricing strategies to determine the appropriate price for a product \cite{PricingStrategiesCorporateProfitability,DataPricingMachineLearningPipelines}.  
In this paper, we concentrate on intrinsic data valuation for practical applications. However, tightly linking intrinsic data valuation to the existence of a validation set can be impractical. A universally accepted validation set may not be available, and a specific validation set might not adequately reflect the data distribution for a given learning task \cite{ValidationFreeReplicationRobustVolume}.
Moreover, possessing a validation set can enable malicious sellers to alter their datasets to overfit the validation set. Additionally, focusing on a specific ML model or task for data valuation may not align with the interests of all stakeholders. Therefore, we adopt an intrinsic data valuation approach that does not rely on validation requirements and is performed prior to any tasks such as training an ML model (Amiri et al. 2023).  

As one of the applications, in the rapidly evolving field of personalized medicine, particularly for cancer treatment, oncology researchers aim to identify the most effective therapies tailored to individual patients' genetic profiles. This involves leveraging data marketplaces (Agarwal et al. 2019) to find potential biomarkers and therapeutic targets that match patient-specific datasets. Researchers prioritize biomarkers with proteomic profiles similar to those found in their datasets. These genetic and proteomic interactions are often represented as graphs, where nodes represent genes or proteins, and edges represent their interactions \cite{protein1,protein2,protein3}. Structural similarity in genetic interactions can indicate similar responses to specific therapies, which is crucial for identifying effective treatments or repurposing existing drugs \cite{personalized1}. Conversely, identifying dissimilar genetic structures can be essential to avoid adverse reactions and resistance, particularly when considering off-target effects or treatment for different cancer subtypes \cite{personalized2,personalized3}. Similarly, in drug discovery, structural similarity in molecules can suggest similar biological activity, which is important for repurposing drugs or optimizing lead compounds to improve efficacy and reduce side effects (Zitnik et al. 2018). Conversely, dissimilar structures might be preferred to avoid cross-reactivity and adverse effects, especially when dealing with off-target interactions or developing drugs for different disease subtypes \cite{drugoff}.  Traditional approaches \cite{datavalue2,datavaluereview,datavalue1} to graph dataset comparison often rely on feature-based metrics that do not fully capture the intricate structural similarities and differences in the subjected graphs. A significant challenge arises due to the lack of visibility into sellers' graph data. Sending subgraphs is impractical and irrelevant. This lack of direct access makes it challenging to accurately assess the value and relevance of external datasets. Recent work by \cite{yao} addresses the problem of data valuation for graphs using Shapley valuation \cite{DataShapleyEquitableValuation}. However, their approach is not task-agnostic and requires a validation set to compute the utility of the data valuation, which may not be practical. Additionally, a significant drawback of using Shapley valuation is the computational infeasibility; the computational cost grows exponentially with the number of samples, necessitating approximation methods that can compromise performance.

Inspired by the mentioned challenges for valuing graph datasets, our paper introduces three metrics for evaluating a seller's graph dataset for a buyer in a task-agnostic manner, focusing on structural and featural attributes of graphs. We assume a buyer and a seller, each with their own graph datasets, where the goal is to value the seller's graph data for the buyer. We break down each graph into its structural and featural aspects, analyze them separately, and integrate the analysis to understand their unique characteristics. In particular, we use structural attributes to capture the distance between graphs' structures and use their featural attributes to measure the similarities and differences in their statistical properties. We enable this by using a blind message passing framework with two unique characteristics. 
\begin{enumerate} \item \textbf{Double-Blindness:} It ensures double-blindness to each party’s dataset, meaning that the buyer does not have access to part or all of the seller’s dataset, and vice versa. This prevents any party from tampering with or gaining advantage from the data. \item \textbf{Task-Agnostic:} It is task-agnostic, meaning it is not dependent on a specific learning algorithm or utility function. Instead, it takes the output of a learning algorithm (machine learning model) and/or a dataset as input and outputs a real-value score. This makes the framework generalizable to any context that uses graphs as its modality. 
\end{enumerate}  

Overall, our contributions can be summarised as follows:

\begin{itemize}
    \item We introduce a novel metric, termed structural disparity, specifically designed for graph datasets. This metric can be utilized independently or in conjunction with the featural attributes of data to offer a comprehensive valuation of graph datasets.

    \item We introduce a framework called blind message passing for task-agnostic graph dataset exchange. Our proposed framework is adaptable to datasets with varying graph structures, node/edge types, and sizes. It ensures double-blindness, meaning neither the buyer nor the seller has access to the counterpart’s data. This prevents data manipulation and ensures fair valuation.

\end{itemize}

To the best of our knowledge, this is the first paper to investigate data valuation for graphs in a task-agnostic manner, eliminating the need for validation sets. We hope our work will inspire further research in this area, given its broad potential applications across various fields.

\section{Preliminary}

\textbf{Graph representation:} Let a graph $G$ be defined as $G = (\mathcal{V}, \mathcal{E})$ where $\mathcal{V} = \{v_1, \ldots, v_N\}$ is the set of nodes, with a cardinality $|\mathcal{V}| = N$, and $\mathcal{E} = \{e_1, \ldots, e_M\}$ is the set of edges, with a cardinality $|\mathcal{E}| = M$. $G$ can be represented by an adjacency matrix $A \in \{0, 1\}^{N \times N}$, with $A_{ij} = 1$ if nodes $v_i$ and $v_j$ are connected and $A_{ij} = 0$ otherwise.

\noindent \textbf{$L_1$-Wasserstein distance:} The Wasserstein distance is a measure of dissimilarity between probability distributions defined on a specific metric space. Let's denote two such distributions as $ p_1 $ and $ p_2 $, operating on a metric space $\mathcal{H}$ \cite{wass}. The $ L_1 $-Wasserstein distance with Euclidean distance as the ground distance is given by:
\begin{align}
W_1(p_1, p_2) =  \inf_{\gamma \in \Gamma(p_1, p_2)} \int_{\mathcal{H} \times \mathcal{H}} ||x - y||_2 d\gamma(x, y).    
\end{align}
Here, $\Gamma(p_1, p_2)$ denotes the set of all possible couplings (or joint distributions) $\gamma$ whose marginals are $p_1$ and $p_2$, respectively. The term $||x - y||_2$ represents the Euclidean distance between points $x$ and $y$ in the metric space $\mathcal{H}$. The integral $\int_{\mathcal{H} \times \mathcal{H}} ||x - y||_2 d\gamma(x, y)$ computes the expected value of the Euclidean distance between the points under the coupling $\gamma$.

\noindent \textbf{Data marketplace:} In data marketplace we assume that there exists multiple sellers and multiple buyers each with their own graph datasets. The objective is to find the relative value of the sellers' datasets with respect to the datasets that that the buyers already have. For the sake of simplicity, we assume a single buyer and a single seller scenario. We denote the set of graphs in the buyer and seller by $\mathcal{G}^b =\{G^b_1, \ldots, G^b_{n^b}\}$ and $\mathcal{G}^s =\{G^s_1, \ldots, G^s_{n^s}\}$, respectively, where $G^l_i = (\mathcal{V}^l_i, \mathcal{E}^l_i)$ with $|\mathcal{V}^l_i| = N_i^l$ and $|\mathcal{E}^l_i| = M_i^l$. Each graph $G^l_i$ has the adjacency matrix $A^l_i \in \{0, 1\}^{N_i^l \times N^l_i}$, for $l \in \{b, s\}$. Furthermore, the nodes' features of the graph $G^l_i$ is $X^l_i \in \mathbb{R}^{N^l_i \times r}$, where $r$ is the number of features for each graph's node. We define $X^l$ as the vertical concatenation of $X^l_i$, i.e., $X^l := \begin{bmatrix} (X^l_1)^T & \cdots & (X^l_{n^l})^T \end{bmatrix}^T \in \mathbb{R}^{N^l \times r}$, where we define $N^l := \sum_{i=1}^{n^l} N^l_i$. 

\noindent \textbf{Diversity and relevance:} Following the work \cite{datavalue2}, we argue that the featural attributes of the graphs can be effectively represented by two metrics: \textit{diversity} and \textit{relevance}. We employ second moment summary statistics, specifically the empirical covariance matrix, to capture the statistical properties of the features. Next, we present the approach in \cite{datavalue2} in estimating diversity and relevance between the features of nodes\footnotemark in buyer and seller. \footnotetext{We highlight that, in this paper, we employ the features of the graphs' \textit{nodes}, while one can extend the procedure for the features of the graphs' \textit{edges}. } To estimate diversity and relevance, first, the buyer performs eigendecomposition on the covariance matrix \(\frac{1}{N^b} (X_b)^T X_b\); i.e.,
\begin{align}
\frac{1}{N^b} (X^b)^T X^b = U \text{Diag}(\lambda_1, \ldots, \lambda_r) U^T,    
\end{align}

where $\lambda_i$ is the $i$-th largest eigenvalue of $\frac{1}{N^b} (X^b)^T X^b$, and $U = [u_1 \ldots u_r]$ with $u_i \in \mathbb{R}^r$ denoting the eigenvector corresponding to the eigenvalue $\lambda_i$. We note that $\lambda_i \geq 0$ since $\frac{1}{N^b} (X^b)^T X^b$ is a positive semi-definite matrix. We further note that $u_1, \dots, u_r$ represent the principal directions containing the most significant information in the covariance matrix $\frac{1}{N^b} (X^b)^T X^b$. The buyer shares the eigenvectors $u_1, \ldots, u_r$ with the seller, while the eigenvalues $\lambda_1, \ldots, \lambda_r$ stay local at the buyer. The seller then estimates the variance of its covariance matrix $\frac{1}{N^s} (X^s)^TX^s$ along $u_1, \ldots, u_r$, the principal directions important to the buyer. This is carried out as:
\begin{align}
\hat{\lambda}_i = \left\lVert  \frac{1}{N^s} (X^s)^TX^s u_i\right\rVert\ , \quad i = 1, \ldots, r,    
\end{align}
where the covariance matrix $\frac{1}{N^s}(X^s)^TX^s$ is first projected into $u_i$ and then the $\ell_2$-norm of the resultant vector provides the estimate of the variance (the data matrices are zero-centered). We note that if $u_i$ is an eigenvector of $\frac{1}{N^s} (X^s)^TX^s$, then $\hat{\lambda}_i$ is its corresponding eigenvalue. We also note that, intuitively, $\lambda_i$ and $\hat{\lambda}_i$ capture the significance of information with, respectively, the buyer's and seller's data along a principal direction of the buyer's data. Buyer and seller share $\lambda_i$ and $\hat{\lambda}_i$, for $i = 1, \ldots, r$, respectively, with the broker, which uses this information to estimate the diversity and relevance of the seller’s data for the buyer. We estimate diversity and relevance based on the volume of the space specified by the coordinates corresponding to the principal components of the covariance matrix of the buyer’s data:
\begin{align}
D &=  \prod_{i=1}^{r} \left( \frac{|\lambda_i - \hat{\lambda}_i|}{\max\{\lambda_i, \hat{\lambda}_i\}} \right)^{1/r}, \label{eq:13} \\
R &= \prod_{i=1}^{r} \left( \frac{\min\{\lambda_i, \hat{\lambda}_i\}}{\max\{\lambda_i, \hat{\lambda}_i\}} \right)^{1/r}.    
\label{eq:14}
\end{align}
The diversity is correlated with the volume of the difference between the variance of the buyer’s and seller’s data in each coordinate; that is, $\prod_{i=1}^{r} |\lambda_i - \hat{\lambda}_i|$. On the other hand, the relevance is correlated with the volume occupied by both buyer’s and seller’s data in these coordinates; that is,
$\prod_{i=1}^{r} \min\{\lambda_i, \hat{\lambda}_i\}$. Furthermore, we normalize these estimates by dividing them by the entire volume, i.e., $\prod_{i=1}^{r} \max\{\lambda_i, \hat{\lambda}_i\}$. Finally, we use geometric mean to keep these metrics within a reasonable range, particularly in the interval $[0, 1]$. It is easy to verify that $0 \le D+R \le 1$. Given the two metrics presented—diversity and relevance for comparing graphs based on their featural attributes—we will introduce methods for comparing graphs based on their structural attributes in the following section.

\section{Structural attributes}\label{SecSA}

In this section, we present a method to measure the structural disparity between two graphs using their positional and structural embeddings \cite{rwpe, wolff}. To measure the structural disparity between graphs, it is essential to obtain embeddings for graphs that extract rich structural and positional features. We propose using the two common methods to generate these embeddings using positional and structural encodings to embed each graph, irrespective of its feature attributes.

\noindent \textbf{Random walk structural embedding (RWSE)} \cite{rwpe}: RWSEs are defined for \( k \) steps of random walk for node \( i \) of the graph:
\begin{align}
z_i^{RW}= \begin{bmatrix}
RW_{ii} & RW_{ii}^2 & \cdots & RW_{ii}^k
\end{bmatrix}^T  \in \mathbb{R}^{k},    
\end{align}
where $RW_{ii}^j$ is the probability of getting back to node $i$ after $j$ steps when we start walking from node $i$. The random walk operator is defined as \( RW = A B^{-1} \), where \( A \in \mathbb{R}^{N \times N} \) is the adjacency matrix and \( B \in \mathbb{R}^{N \times N} \) is the degree matrix and $RW_{ii}$ is the $i$-th diagonal entry of $RW$.  

\begin{figure*}
\centering
\includegraphics[width=0.9\textwidth]{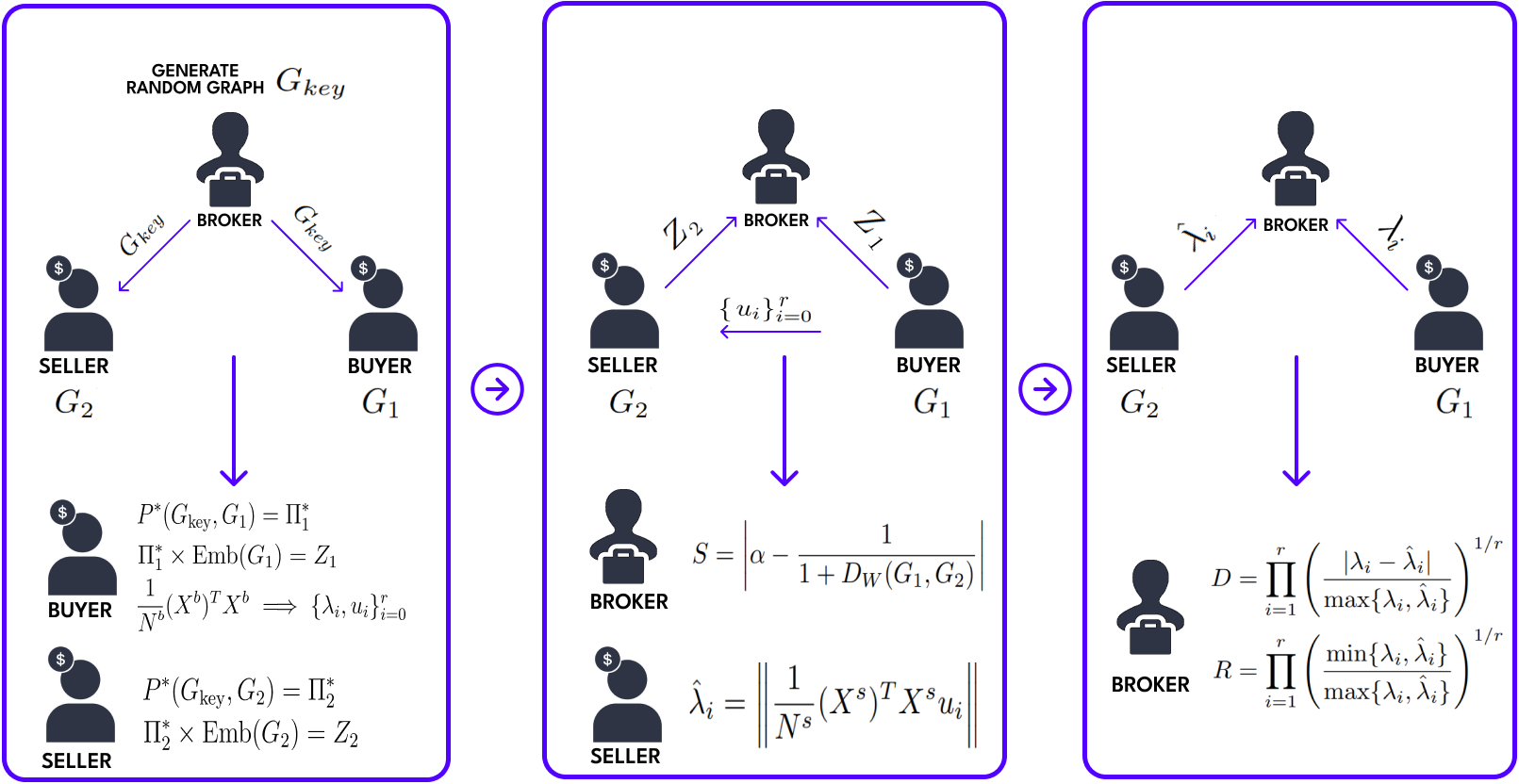}
\caption{\label{fig:frog} The BMP framework for task-agnostic graph data valuation involves three steps: (Left) A trusted broker generates a random proxy graph and shares it with the buyer and seller, who then compute optimal permutations and embeddings. The buyer performs eigendecomposition on the covariance of her feature matrix to find eigenvalues and eigenvectors. (Middle) The buyer and seller send their embeddings to the broker, who computes the structural disparity \(S\). (Right) The buyer and seller share their eigenvalues with the broker, who computes relevance \(R\) and diversity \(D\).}
\end{figure*}

\noindent \textbf{Laplacian eigenvenctor positional embedding (LapPE)} \cite{lapPE}: LapPE provides accurate embeddings of graphs into Euclidean spaces.\ It is constructed by factorizing the graph Laplacian, defined as \(\Delta = I_N - B^{-1/2}AB^{-1/2} = U\Lambda U^T\), where \( I_N \) is the \( N \times N \) identity matrix, and the matrices \( \Lambda \) and \( U \) represent the eigenvalues and eigenvectors, respectively. The absolute value of $\ell_2$-norm of the first non-trivial \( k'\) eigenvectors for node \( i \) is denoted as \( LP_{i}\). Hence, we define LapPE as:
\begin{align}
z_i^{LP} = \begin{bmatrix}
LP_{i1} & LP_{i2} & \cdots & LP_{ik'}
\end{bmatrix}^T \in \mathbb{R}^{k'}.    
\end{align}
To create an expressive embedder, we concatenate the RWSE and LapPE embeddings for each node:
\begin{align}
z_i^{PE} = \text{concatenate}(z_i^{RW}, z_i^{LP}) \in \mathbb{R}^{k+k'}.
\end{align}
Finally, the overall embedding for the graph is constructed by concatenating the positional and structural encodings of all nodes:
\begin{align}
Z = \begin{bmatrix}
z_1^{PE} & z_2^{PE} & \cdots & z_N^{PE}
\end{bmatrix} \in \mathbb{R}^{(k+k') \times N},
\end{align}
Accordingly, we define function $\text{Emb}(\cdot):G \to \mathbb{R}^{(k+k') \times N}$, which takes a graph and outputs the embedding $Z$.

Our main objective is to develop a method for comparing graphs based on their structural properties. We propose using the graph Wasserstein distance (GWD), inspired by \cite{wwl}. For simplicity, consider two graphs $G_1= (\mathcal{V}_1, \mathcal{E}_1)$ and $G_2= (\mathcal{V}_2, \mathcal{E}_2)$; our goal is to compute GWD between them. To this end, we first need to align the two graphs; that is, we need to find a consistent permutation between the underlying graphs. For this purpose, in the following, we define graph matching on matrices.

\noindent \textbf{Definition (Graph Matching\footnotemark): } Given two graphs \( G_i = (\mathcal{V}_i, \mathcal{E}_i) \) and their normalized Laplacian matrices \( L_i \) for \( i \in \{1, 2\} \), their matching can be represented by a permutation matrix \( P \in \Pi \) that optimally aligns the graph structures. Formally, the optimal permutation \( P^* \) is obtained as \footnotetext{The graph matching can be extended to matching two graphs with unequal numbers of nodes by padding the Laplacian matrix of one graph with zeros.}:
\begin{align}\label{EQGraphMatching}
P^*(G_1, G_2) = \arg \min_{P \in \Pi} \left\| L_1 - P^T L_2 P \right\|_F.     
\end{align} 

\noindent \textbf{Definition (Distance-Compatible):} Distance-compatible permutation of two graphs \( G_i = (\mathcal{V}_i, \mathcal{E}_i) \) is a permutation set \((\Pi_1^*, \Pi_2^*)\) such that:

\begin{align}\label{EqDCTwoGraphs}
P^*(G_1,G_2) = \Pi_1^*, \quad P^*(G_2,G_1) = \Pi_2^*.
\end{align}

\begin{figure}[t!]
\centering
\includegraphics[width=0.45\textwidth]{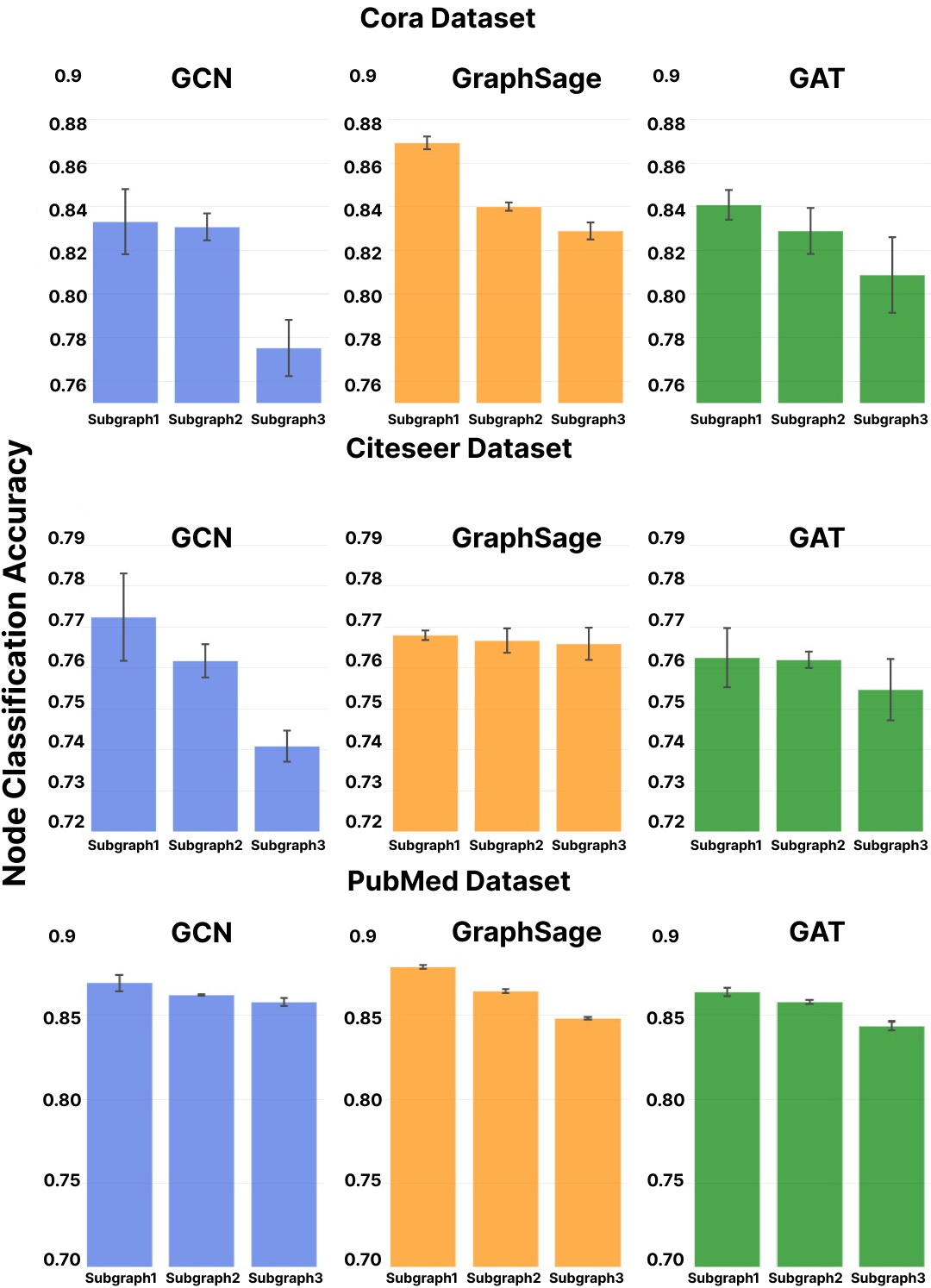}
\caption{\label{fig:chart} Node classification accuracy of datasets prodivded in Table \ref{tab:datasets} via subset selection using the BMP framework.}
\end{figure}

\noindent \textbf{Definition (Graph Wasserstein Distance):} Given two distance-compatible graphs $G_1 = (\mathcal{V}_1,\mathcal{E}_1)$ and $G_2 = (\mathcal{V}_2,\mathcal{E}_2)$ and $|\mathcal{V}| = \max \{|\mathcal{V}_1|, |\mathcal{V}_2|\}$ with respective permutation set $(\Pi_1^*, \Pi_2^*)$ and $\text{Emb}(G_l) : G_l \to \mathbb{R}^{(k+k') \times |\mathcal{V}_l|}$ as the embedder we define GWD as\footnote{We pad one of the graphs with zeros in order to be able to sum the two of them with size mismatch}: 
\begin{align}
Z_l &= \Pi_l^* \times \text{Emb}(G_l), \quad l \in \{1,2\},\\
D_W(G_1, G_2) &:= \sum^{|\mathcal{V}|}_{i=1} W_1(Z_1[:, i], Z_2[:, i]),   \label{eq:8}
\end{align}
where $Z_l[:, i]$ is the $i$-column of $Z_l$.

In the above approach, we embed the graphs, focusing solely on their structure. Prior works \cite{pos1}, \cite{pos2} have demonstrated that positional embedders can effectively capture the structural characteristics of graphs. After embedding the graphs with positional embedders, we interpret the resulting embeddings as empirical distributions and use the Wasserstein distance to compare these distributions. To compute the GWD, we find the pairwise Wasserstein distances between corresponding nodes and sum these distances for distance-compatible graphs. It is important to note that each column $Z_l[:, i]$ in \eqref{eq:8} represents a realization sampled from an underlying unknown probability distribution. Consequently, our objective involves computing the $W_1$ distance between these distributions. Next, we use the GWD notion to define the structural disparity between two graphs, which will be used for data valuation between buyer's and seller's graph datasets.

\noindent \textbf{Definition (Structural Disparity):} The structural disparity metric $S$ between two graphs $G_1$ and $G_2$ is defined as:
\begin{align}\label{Structural_Disparity_eq}
S = \left|\alpha - \frac{1}{1 + D_W(G_1, G_2)}\right|, \quad \alpha \in [0,1],    
\end{align}
where $D_W(G_1, G_2)$ is the GWD between $G_1$ and $G_2$. The parameter $\alpha$ represents the preference for the amount of disparity: $\alpha = 1$ indicates a preference for greater differences between the graphs, while $\alpha = 0$ indicates a preference for greater similarity. For identical graphs, \( D_W(G_1, G_2) \) would be $0$, and \( S \) would equal \( |\alpha - 1| \). With \( \alpha = 1 \) (indicating a preference for high differences), \( S \) would be 0. Conversely, with \( \alpha = 0 \) (indicating a preference for low differences), \( S \) would be at its maximum value of 1. In the extreme case of infinite GWD between \( G_1 \) and \( G_2 \), \( S \) would be \( |\alpha| \). With \( \alpha = 1 \), \( S \) would reach its maximum, and with \( \alpha = 0 \), it would be at its minimum.

\section{Blind message passing}\label{SecBMP}

In this section, we build on the concepts introduced earlier to compare the structural attributes of buyer and seller graphs within a data marketplace.\ This requires addressing two key challenges.\ First, both the buyer and seller possess multiple graphs, necessitating an extension of the structural disparity measure—originally designed for pairs of graphs—to accommodate comparisons between two sets of graphs.\ Second, it is crucial to ensure that the graphs remain local and are not shared.\ We further demonstrate how the disentanglement works in practice through the blind message passing (BMP) framework for the case both parties having multiple graphs.

One of the most important aspects of this framework lies in the blind exchange of data between both parties—neither the seller nor the buyer has access to the counterpart's data, preventing any party from tampering with its own data to gain an advantage. As depicted in Fig. \ref{fig:frog}, a trustworthy broker who initially generates a simple proxy graph from some distribution, denoted by $G_{key}$ which remains unknown to both parties. To preserve privacy and distribute the computational cost among parties, $G_{key}$ is shared with both the seller and the buyer. Both parties then find their respective optimal permutations with regard to the proxy graph $G_{key}$ using the graph matching technique in Eq. \eqref{EQGraphMatching}. Next, we define $\varepsilon$-Conformity, which will be used to calculate the error in finding the optimal permutation using the proxy graph.

\noindent \textbf{Definition ($\varepsilon$-Conformity):} Two graphs $G_1$ and $G_2$ with permutations $\Pi_1$ and $\Pi_2$ are defined as $\varepsilon$-conform if:
\begin{align}
\left\| \Pi_1^T L_1 \Pi_1 - \Pi_2^T L_2 \Pi_2 \right\|_F \leq \varepsilon.    
\end{align}

\noindent \textbf{Corollary 1 (Transitivity):} Two graphs $G_1$ and $G_2$ that are both matched with $G_{key}$ are $\hat{\varepsilon}$-conform with respect to $\Pi_1^* \triangleq P^*(G_{key}, G_1)$ and $\Pi_2^* \triangleq P^*(G_{key}, G_2)$ for
\begin{align}
\hat{\varepsilon} = \left\| L_{key} - {\Pi_1^*}^T L_1 \Pi_1^* \right\|_F + \left\| L_{key} - {\Pi_2^*}^T L_2 \Pi_2^* \right\|_F,
\end{align}
where $L_{key}$, $L_1$, and $L_2$ are the normalized Laplacian for $G_{key}$, $G_1$, and $G_2$, respectively. Proof provided in the supplementary materials.

\noindent \textbf{Remark:} The transitivity corollary ensures that the upper bound on the error when calculating the optimal permutation of $G_1$ and $G_2$ indirectly using a proxy graph like $G_{key}$ is $\hat{\varepsilon}$. We performed an experiment to demonstrate that incorporating graph matching yields competitive results compared to approaches that do not utilize a proxy graph, thereby validating the practicality of the theoretical upper bound. Details are provided in the supplementary materials.

\begin{table}[t!]
\centering
\caption{Statistics of Datasets.}
\label{tab:datasets}
\begin{adjustbox}{width=0.48\textwidth}
\begin{tabular}{lcccc}
\toprule
\textbf{Dataset} & \textbf{Graph\#} & \textbf{Class\#} & \textbf{Avg Node\#} & \textbf{Avg Edge\#} \\
\midrule
PubMed           & 1  & 3 & 19717  & 44338  \\
Citeseer         & 1  & 6 & 3312  & 4732  \\
Cora   & 1  & 7 & 2708  & 5429  \\
DD           & 1178  & 2 & 284.32 & 715.66 \\
AIDS    & 2000 & 2 & 15.69  & 16.20  \\
PROTEINS       & 1113  & 2 & 39.06  & 72.82  \\
MUTAG          & 188   & 2 & 17.93  & 19.79  \\
DHFR          & 756  & 2 & 42.43  & 44.54  \\
ENZYMES        & 600   & 6 & 32.63  & 62.14  \\
\bottomrule
\end{tabular}
\end{adjustbox}
\end{table}

The process of matching the graphs in both parties with $G_{key}$ aligns the graphs of both parties to a near-distance-compatible node permutation, enabling them to compare structural differences. After computing node embeddings for all the graphs $G_{i}^b (\mathcal{V}_i^b, \mathcal{E}_i^b) \in \mathcal{G}^b$ using positional embedders with the permutation obtained from $P^*(G_{key}, G^b_i)$, $\forall i \in \{1, ..., n^b\}$, the result is a tensor $\Phi^b \in \mathbb{R}^{|\mathcal{G}^b| \times |\mathcal{V}^b| \times (k+k')}$ for the buyer, where $|\mathcal{V}^b| = \max_i \{|\mathcal{V}^b_i|\}$, $(k+k')$ is the dimension of the embedding, and we zero-pad all the embeddings to have a dimension $|\mathcal{V}^b| \times (k+k')$. We note that the embeddings for each node of each graph are of dimension $(k+k')$. Similarly, we can obtain the tensor $\Phi^s \in \mathbb{R}^{|\mathcal{G}^s| \times |\mathcal{V}^s| \times (k+k')}$ for the seller. We now define the mean-pool on the tensor $\Phi^l$, $l \in \{b, s\}$, as follows:

\begin{table*}[t]
\centering
\caption{Graph classification accuracy with a standard deviation of GCN on ranked sets across different datasets, ranked based on the average ranking of $(D)$, $(R)$, and $(S)$ for the train samples. Accuracy results are averaged over 4 random seeds. The top results for each dataset are highlighted in bold.}
\begin{adjustbox}{width=1\textwidth}
\begin{tabular}{@{}lccccc@{}}
\toprule
\textbf{Dataset} & \textbf{Set 1} & \textbf{Set 2} & \textbf{Set 3} & \textbf{Set 4} & \textbf{Set 5} \\ \midrule
DD        & \textbf{0.7063 ± 0.0011} & 0.6851 ± 0.0038 & 0.6885 ± 0.0009 & 0.6885 ± 0.0012 & 0.6978 ± 0.0092 \\ \midrule
MUTAG     & \textbf{0.6158 ± 0.0004} & 0.6053 ± 0.0011 & 0.6105 ± 0.0009 & 0.6053 ± 0.0012 & 0.5842 ± 0.0092 \\ \midrule
AIDS      & 0.7955 ± 0.0004 & \textbf{0.8080 ± 0.0011} & 0.8080 ± 0.0032 & 0.7921 ± 0.0001 & 0.7778 ± 0.0001 \\ \midrule
Proteins  & \textbf{0.6466 ± 0.0005} & 0.6228 ± 0.0003 & 0.5615 ± 0.0008 & 0.5858 ± 0.0013 & 0.6324 ± 0.0010 \\ \midrule
DHFR      & \textbf{0.6232 ± 0.0002} & 0.5953 ± 0.0002 & 0.6053 ± 0.0002 & 0.5879 ± 0.0001 & 0.5174 ± 0.0001 \\ \midrule
ENZYMES   & 0.6250 ± 0.0354 & \textbf{0.6333 ± 0.0102} & 0.6166 ± 0.0522 & 0.6250 ± 0.0154 & 0.6250 ± 0.0154 \\ \bottomrule
\end{tabular}
\end{adjustbox}
\label{tab:performance}
\end{table*}

\noindent \textbf{Definition (Tensor Mean-Pool):} For a tensor $\Phi^l \in \mathbb{R}^{|\mathcal{G}^l| \times |\mathcal{V}^l| \times (k+k')}$, where $|\mathcal{V}^l| = \max_i \{|\mathcal{V}_i^l|\}$ for the graph $G_i^b(\mathcal{V}_i^b, \mathcal{E}_i^b) \in \mathcal{G}^l$ and $(k+k')$ is the dimension of the embedding, we define a mean-pool function $f$ such that $f: \mathbb{R}^{|\mathcal{G}^l| \times |\mathcal{V}^l| \times (k+k')} \to \mathbb{R}^{|\mathcal{G}^l| \times |\mathcal{V}^l|}$:  
\begin{align}
f(\Phi^l) = \frac{1}{(k+k')} \sum_{i=1}^{(k+k')} \Phi^l[:, :, i], \quad l \in \{b, s\},    
\end{align}
where $\Phi^l[:, :, i] \in \mathbb{R}^{|\mathcal{G}^l| \times |\mathcal{V}^l|}$ is a matrix including all the entries in the first and second dimensions of the three-dimensional tensor $\Phi^l$ corresponding to the $i$-th entry of its third dimension. We note that $f(\cdot)$, as defined above, provides a first order summary statistics along the embeddings.

Next, we present our methodology to obtain the structural disparity, as well as the relevance and diversity based on the buyer's and seller's graphs $\mathcal{G}^b$ and $\mathcal{G}^s$, respectively.  After performing mean-pooling, both parties transmit their respective matrices $f(\Phi^b)$ and $f(\Phi^s)$ to the broker for calculating the Wasserstein distance, which measures their structural disparity. Let's denote $ f(\Phi^l) = \begin{bmatrix}
f^l_1 & f^l_2 & \cdots & f^l_{|\mathcal{V}^l|}
\end{bmatrix} $ where $f^l_i \in \mathbb{R}^{|\mathcal{G}^l|}$, $l \in \{b, s\}$. Next, by abusing the notation, GWD is obtained for the sets of graphs $\mathcal{G}^b$ and $\mathcal{G}^s$ according to
\begin{align}
D_W(\mathcal{G}^b, \mathcal{G}^s) = \sum^{|\mathcal{V}'|}_{i=1} W_1(f^b_i, f^s_i)    
\end{align}
which follows from Eq. \eqref{eq:8} where $|\mathcal{V}'| = \max\{|\mathcal{V}^b|, |\mathcal{V}^s|\}$. We then plug the GWD $D_W(\mathcal{G}^b, \mathcal{G}^s)$ into Eq. \eqref{Structural_Disparity_eq} to obtain the structural disparity, i.e.,
\begin{align}\label{EqSDGeneralCase}
S = \left|\alpha - \frac{1}{1 + D_W(\mathcal{G}^b, \mathcal{G}^s)}\right|, \quad \alpha \in [0,1].     
\end{align}

\noindent \textbf{Disentanglement in Practice:} We now turn to the practical implementation of structural and featural disentanglement for graph data valuation. Each of the three metrics diversity $(D)$, relevance $(R)$, and structural disparity $(S)$, as defined in \eqref{eq:13}, \eqref{eq:14}, and \eqref{EqSDGeneralCase}, respectively, offers unique insights into the dataset and can be used independently based on user preferences.\ These metrics are versatile and can be integrated into various utility functions tailored to specific contexts or datasets.\ To ensure generalizability, we propose using an average ranking approach.\ Specifically, consider a scenario where multiple sellers each possess a set of graphs, and there is a single buyer with their own graph set.\ We compute $(D)$, $(R)$, and $(S)$ for each seller's graph set relative to the buyer's graph set. Each seller's set is then ranked according to these three metrics.\ The final ranking and valuation of the sellers' graph sets are obtained by averaging their rankings across all three metrics.\ As demonstrated in our experiments, this average ranking approach is effective in practice and suitable for real-world applications.

\noindent \textbf{Computational Complexity:}  Our framework consists of three primary computationally intensive algorithms. The first algorithm focuses on computing GWD. The naive approach to computing the Wasserstein distance has a complexity of \(O(n^3 \log(n))\), where \(n\) denotes the number of node embeddings or the number of nodes in the graphs. To mitigate this computational burden, we employ several efficient acceleration techniques. Notably, approximations based on Sinkhorn regularization have been proposed \cite{wass2}, which can significantly reduce the complexity to near-linear time.  

The second algorithm addresses the graph matching problem via the linear assignment problem. To optimize computational efficiency and minimize the load on the broker, we use a proxy graph $G_{\text{key}}$. Instead of transmitting both datasets directly to the broker for the computation of graph matching, we delegate this task to the buyer and seller. By offloading the graph matching computation to the buyer and seller, the broker's responsibility is reduced to solely computing GWD.
For two graphs $G_1 = (\mathcal{V}_1, \mathcal{E}_1)$ and $G_2 = (\mathcal{V}_2, \mathcal{E}_2)$ with $|\mathcal{V}_1| =N_1$ and $|\mathcal{V}_2| =N_2$ and $N=\max \{N_1, N_2\}$, the paper \cite{blind} shows that the solution to the graph matching problem can be approximated by the following linear assignment problem and can be efficiently solved by the Hungarian method \cite{hungarian}:

\begin{align}
P^{**} = \arg \max_{P \in \Pi} \text{tr}(P^T \bar{U}_1 (\bar{U}_2)^T).
\end{align}
Here  $U_i \in \mathbb{R}^{N \times N}$ is the orthogonal matrix corresponding to eigenvectors from the eigendecomposition of $L_i = U_i \Lambda_i U_i^T$ and $\bar{U_i}$ is the matrix containing the absolute value of the entries of  $U_i$, for $i \in \{1, 2\}$. The complexity of the Hungarian algorithm is \(O(n^3)\), where \(n\) is the number of nodes or, equivalently, the dimensions of the permutation matrix.

The third algorithm involves the eigendecomposition of the covariance matrix. The eigendecomposition of an \(n \times n\) matrix using standard numerical methods, such as the QR algorithm, typically has a computational complexity of \(O(n^3)\). Considering these components, the overall complexity of our framework is \(O(n^3)\) and \(n\) represents the number of nodes.

\section{Experiments}

We evaluate the BMP framework on four tasks: (i) dataset scoring using structural disparity on node level prediction then evaluating various graph neural network (GNN) models on these datasets, (ii) dataset scoring using relevance, diversity, and structural disparity and then evaluating graph convolutional network (GCN) \cite{gcn} model on these datasets, (iii) assessing if the structural disparity metric can distinguish between different contexts, and (iv) testing the practical performance of the featural metrics. We used datasets from Table \ref{tab:datasets} for the experiments. Details about experiments (iii), (iv), and the experimental setup can be found in the supplementary materials.

In the first task, we explore whether structural disparity can effectively identify the most suitable graph among three candidates offered by sellers, given a baseline graph from the buyer. To quantify each graph's value, we augment the buyer's graph with the seller's graph and evaluate performance on a test set. We start by embedding a large graph using positional embedders, then partition it into three sets: buyer, seller, and test, with the test set containing 20\% of the nodes and the buyer set 10\%. Structural disparity serves as the sole metric for this task. We first train a simple GCN on the seller nodes to generate unsupervised node embeddings. These embeddings are then clustered using K-Means \cite{kmeans}, resulting in three distinct candidate sets. We use the BMP framework to assess the structural disparity between each candidate set and the buyer's set. The candidate sets are ranked by their proximity to the buyer's set, with the candidate set 1 denoted by subgraph 1 in Fig. \ref{fig:chart} being the most similar. We then evaluate the performance using three different GNN models: GCN (distinct from the one used for clustering), GraphSAGE \cite{sage} with a mean aggregator, and GAT \cite{gat} with an attention mechanism. For each GNN, we train three models, each using a different candidate set combined with the baseline set. The test mask is used for evaluating node classification accuracy. As summarized in Fig. \ref{fig:chart}, our results demonstrate a clear trend: lower structural disparity between a candidate set and the baseline correlates with higher node classification accuracy. Clustering was employed to ensure distinction between candidates.

In the second task, we evaluate our approach for capturing both structural and featural attributes between graphs using three metrics: $(D)$, $(R)$, and $(S)$. The dataset is divided into three disjoint sets: baseline, training, and test, with the test set comprising 20\% of the data and the baseline 10\%. Recognizing that relying on a single aggregating function from the three metrics could introduce bias dependent on the dataset, we adopt a more general approach. We first rank the dataset samples according to each metric independently, then calculate an average ranking across all metrics, setting $\alpha=0.5$ for this purpose. To assess the relative ranking of each set relative to the baseline, we compute the scores between the baseline and each individual sample in the training set, ranking them from highest to lowest. These ranked samples are then divided into five sets based on their descending rankings (i.e., Set 1 \(\geq\) Set 2, etc.), and each set is extended with the baseline. For this task, we utilized a GCN to perform the graph classification task, training the model on each of these five sets and evaluating its accuracy on the test set. The results are summarized in Table \ref{tab:performance}. The consistently superior performance of the top two sets supports the validity of our three metrics—relevance, diversity, and structural disparity—in evaluating and scoring graph datasets. 
\section{Conclusion}
In this study, we introduced a novel framework for task-agnostic graph data valuation, leveraging both structural and featural representations. Our approach uses blind message passing (BMP) and graph Wasserstein distance (GWD) for effective alignment and comparison of graph structures, ensuring privacy and efficiency in data marketplaces. We demonstrated that our metrics—structural disparity, diversity, and relevance—are effective in capturing the essential characteristics of graph data that are crucial for valuation. Experimental results on real-world datasets validated our method, showing that higher-scored sets lead to improved performance in various applications. This work has significant implications for data marketplaces, enabling accurate data valuation. Future work will extend this framework to more complex graph structures and additional domains.

\bibliography{aaai25}

\clearpage


\appendix

\section{Appendix / Supplementary Materials}

\subsection{Algorithms} \label{algorithms}

Here we outline algorithms for obtaining GWD, structural disparity, diversity, and relevance. 

We note that the buyer’s set of graphs is $\mathcal{G}^b = \{G^b_1, \ldots, G^b_{n^b}\}$, and the seller’s set of graphs is $\mathcal{G}^s = \{G^s_1, \ldots, G^s_{n^s}\}$ with $X^b \in \mathbb{R}^{N^b \times r}$ and $X^s \in \mathbb{R}^{N^s \times r}$ as the features of the graphs' nodes for the buyer and seller, respectively. First, we outline the algorithm for obtaining GWD between the buyer's and seller's graphs:

\begin{algorithm}[H]
\caption{GWD}
\begin{algorithmic}[1]
\STATE \textbf{Input:} $f(\Phi^b) = \begin{bmatrix}
f^b_1 & f^b_2 & \cdots & f^b_{|\mathcal{V}^b|}
\end{bmatrix} \in \mathbb{R}^{|\mathcal{G}^b| \times |\mathcal{V}^b|}$, $f(\Phi^s) = \begin{bmatrix}
f^s_1 & f^s_2 & \cdots & f^s_{|\mathcal{V}^s|}
\end{bmatrix} \in \mathbb{R}^{|\mathcal{G}^s| \times |\mathcal{V}^s|}$, $|\mathcal{V}'| = \max\{|\mathcal{V}^b|, |\mathcal{V}^s|\}$ where $|\mathcal{V}^l| = \max(|\mathcal{V}^l_1|, \ldots, |\mathcal{V}^l_{n^l}|)$, $l \in \{b, s\}$

\IF{ $|\mathcal{V}^b| > |\mathcal{V}^s|$}
    \STATE Zero-pad $f(\Phi^s)$ until it becomes $f(\Phi^s) \in \mathbb{R}^{|\mathcal{G}^s| \times |\mathcal{V}^b|}$
\ELSE{}
    \STATE Zero-pad $f(\Phi^b)$ until it becomes $f(\Phi^b) \in \mathbb{R}^{|\mathcal{G}^b| \times |\mathcal{V}^s|}$

\ENDIF

    \STATE $D_W(f(\Phi^b), f(\Phi^s)) = \sum^{|\mathcal{V}'|}_{i=1} W_1(f^b_i, f^s_i)$ 

\STATE \textbf{Return}: $D_W(f(\Phi^b), f(\Phi^s))$
\end{algorithmic}
\end{algorithm}

\noindent Here we present the algorithm for obtaining $S$, the structural disparity between the buyer's and seller's graphs:

\begin{algorithm}[H]
\caption{Structural disparity ($S$)}
\begin{algorithmic}[1]
\STATE \textbf{Input:} Buyer’s graphs set $\mathcal{G}^b$, seller’s graphs set $\mathcal{G}^s$

\STATE \textbf{Broker}:  

\indent Generate the proxy graph $G_{key}$  

\indent Share $G_{key}$ with the buyer and seller

\FOR{each party $l \in \{{b, s}\}$}

    \FOR{each graph $G^l_i \in \mathcal{G}^l$}
        \STATE Compute the optimal permutation $\Pi^l_i = P^*(G_{key}, G^l_i)$
        \STATE $\Phi^l[i, :, :] = \Pi^l_i \times (\text{Emb}(G^l_i))^T$, where $\text{Emb}(\cdot):G \to \mathbb{R}^{d \times N}$
    \ENDFOR
    \STATE Compute mean-pool matrix $f(\Phi^l) = \frac{1}{d} \sum_{i=1}^{d} \Phi^l[:, :, i]$
    \STATE Share $f(\Phi^l)$ with the broker
\ENDFOR

\STATE \textbf{Broker}:

\indent Compute $D_W(\mathcal{G}^b, \mathcal{G}^s)$

\indent Compute $S = |\alpha - \frac{1}{1 + D_W(\mathcal{G}^b, \mathcal{G}^s)}|$

\indent \textbf{Return} Structural disparity ($S$)

\end{algorithmic}
\end{algorithm}

\noindent Next we outline the algorithm for obtaining diversity $D$ and relevance $R$:

\begin{algorithm}[H]
\caption{Diversity ($D$) and relevance ($R$)}
\begin{algorithmic}[1]
\STATE \textbf{Input:} Buyer’s node features ${X}^b \in \mathbb{R}^{N^b \times r}$, seller’s node features ${X}^s \in \mathbb{R}^{N^s \times r}$

\STATE \textbf{Buyer}:

    \indent Eigendecompose the covariance matrix $\frac{1}{N^b}(X^b)^TX^b = U \text{Diag}(\lambda_1, \ldots, \lambda_r) U^T$  
    
    \indent Share $U = [u_1 \ldots u_r]$ with the {seller}  
    
    \indent Share $\{\lambda_i\}^r_{i=1}$ with the {broker}

\STATE \textbf{Seller}:

    \indent Calculate $\hat{\lambda}_i = \left\lVert  \frac{1}{N^s} (X^s)^TX^s u_i\right\rVert\ $, for $i=1, \dots, r$

    \indent Share $\{\hat{\lambda}_i\}^r_{i=1}$ with the {broker}

\STATE \textbf{Broker}:

    \indent $D =  \prod_{i=1}^{r} \left( \frac{|\lambda_i - \hat{\lambda}_i|}{\max\{\lambda_i, \hat{\lambda}_i\}} \right)^{1/r}$

    \indent $R = \prod_{i=1}^{r} \left( \frac{\min\{\lambda_i, \hat{\lambda}_i\}}{\max\{\lambda_i, \hat{\lambda}_i\}} \right)^{1/r}$  
    
\STATE \textbf{Return}: $D$ and $R$
\end{algorithmic}
\end{algorithm}

\subsection{Proofs} \label{proof}

\noindent \textbf{Corollary 1 (Transitivity):} Two graphs $G_1$ and $G_2$ that are both matched with $G_{key}$ are $\hat{\varepsilon}$-conform with respect to $\Pi_1^* \triangleq P^*(G_{key}, G_1)$ and $\Pi_2^* \triangleq P^*(G_{key}, G_2)$ for
\begin{align}
\hat{\varepsilon} = \left\| L_{key} - {\Pi_1^*}^T L_1 \Pi_1^* \right\|_F + \left\| L_{key} - {\Pi_2^*}^T L_2 \Pi_2^* \right\|_F.
\end{align}
where $L_{key}$, $L_1$, and $L_2$ are the normalized Laplacian for $G_{key}$, $G_1$, and $G_2$, respectively.

\noindent
\textbf{Proof.} Assume that graphs $G_1$ and $G_2$ match with $G_\text{key}$ under permutations $\Pi^*_1$ and $\Pi^*_2$, respectively. We would like to show that $G_1$ and $G_2$ are $\hat{\epsilon}$-conform; that is:
\begin{equation}
\| \Pi^*_1{}^\top L_1 \Pi^*_1 - \Pi^*_2{}^\top L_2 \Pi^*_2 \|_F \leq \hat{\epsilon}. \tag{21}
\end{equation}

\noindent
For the left side of the above inequality we have
\newcommand\firstineq{\mathrel{\overset{\makebox[0pt]{\mbox{\normalfont\tiny\sffamily (a)}}}{\le}}}
\begin{small}
\begin{align}
    & \| \Pi_1^*{^\top} L_1 \Pi^*_1 - \Pi_2^*{^\top} L_2 \Pi^*_2 \|_F \nonumber \\
    &= \| \Pi_1^*{^\top} L_1 \Pi^*_1 - L_{\text{key}} + L_{\text{key}} - \Pi_2^*{^\top} L_2 \Pi^*_2 \|_F \nonumber \\
    &\firstineq \| L_{\text{key}} - \Pi_1^*{^\top} L_1 \Pi^*_1 \|_F + \| L_{\text{key}} - \Pi_2^*{^\top} L_2 \Pi^*_2 \|_F  \nonumber \\
    &= \hat{\epsilon}.
\end{align}
\end{small}

\noindent
where (a) results from the triangle inequality. This completes the proof.

\section{Experimental details}

\subsection{Datasets}
We test 26 commonly used benchmark datasets in our experiments. Except for the first three and last two, all datasets are selected from TUDataset \cite{TU}. These datasets include PubMed, Citeseer, Cora, BZR, COX2, DHFR, MUTAG, ENZYMES, KKI, $\text{Peking\_1}$, PROTEINS, OHSU, $\text{MSRC\_21}$, COIL-DEL, Letter-high, Letter-low, IMDB-BINARY, IMDB-MULTI, twitch-egos, COLORS-3, SYNTHETIC, FRANKENSTEIN, DD, AIDS, MNIST, and CIFAR10. The first three datasets are selected from the Planetoid dataset \cite{planetoid}, and the last two datasets were selected from GNNBenchmark \cite{GNNBenchmark}. The statistics of the datasets that we used in the appendix are summarized in Table \ref{tab:datasets2}.

\begin{table}[t!]
\centering
\caption{Statistics of Datasets.}
\label{tab:datasets2}
\begin{adjustbox}{width=0.48\textwidth}
\begin{tabular}{lcccc}
\toprule
\textbf{Dataset} & \textbf{Graph\#} & \textbf{Class\#} & \textbf{Avg Node\#} & \textbf{Avg Edge\#} \\
\midrule
PubMed           & 1  & 3 & 19717  & 44338  \\
Citeseer         & 1  & 6 & 3312  & 4732  \\
Cora   & 1  & 7 & 2708  & 5429  \\
DD           & 1178  & 2 & 284.32 & 715.66 \\
AIDS    & 2000 & 2 & 15.69  & 16.20  \\
PROTEINS       & 1113  & 2 & 39.06  & 72.82  \\
MUTAG          & 188   & 2 & 17.93  & 19.79  \\
DHFR          & 756  & 2 & 42.43  & 44.54  \\
ENZYMES        & 600   & 6 & 32.63  & 62.14  \\
BZR           & 405  & 2 & 35.75  & 38.36  \\
COX2         & 467  & 2 & 41.22  & 43.45  \\
FRANKENSTEIN   & 4337  & 2 & 16.90  & 17.88  \\
KKI    & 83 & 2 & 26.96  & 48.42  \\
Peking    & 85 & 2 & 39.31  & 77.35  \\
OHSU    & 79 & 2 & 82.01  & 199.66  \\
MSRC\_21    & 563 & 20 & 77.52  & 198.32  \\
COIL-DEL    & 3900 & 100 & 21.54  & 54.24  \\
Letter-high    & 2250 & 15 & 4.67  & 4.50  \\
Letter-low    & 2250 & 15 & 4.68  & 3.13  \\
IMDB-BINARY    & 1000 & 2 & 19.77  & 96.53  \\
IMDB-MULTI    & 1500 & 3 & 13.00  & 65.94  \\
twitch-egos    & 127094 & 2 & 29.67	  & 86.59 \\
COLORS-3     & 10500 & 11 & 61.31  & 91.03  \\
SYNTHETIC    & 300 & 2 & 100.00  & 196.00  \\
MNIST    & 70000 & 10 & 70.6  & 564.5  \\
CIFAR10    & 60000 & 10 & 117.6  & 941.2  \\
\bottomrule
\end{tabular}
\end{adjustbox}
\end{table}

\begin{figure*}[t!]
\centering
\includegraphics[width=1\textwidth]{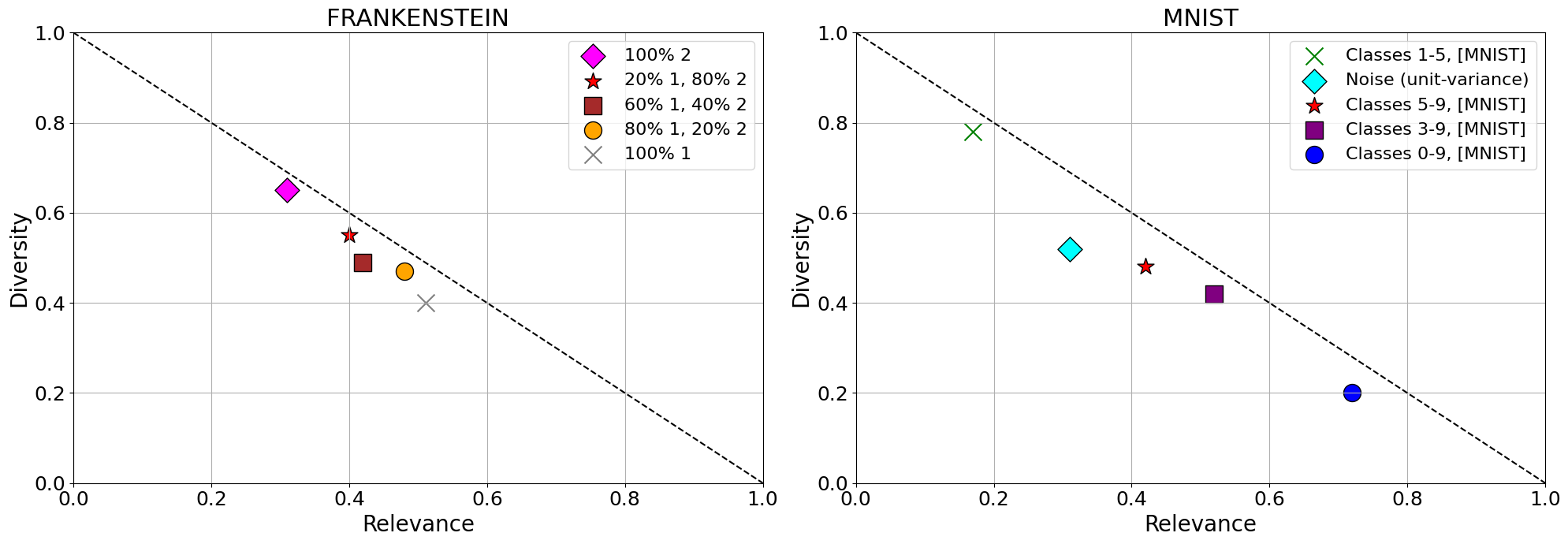}
\caption{\label{fig:div} Estimation of diversity and relevance for FRANKENSTEIN dataset (left) and MNIST dataset (right) }
\end{figure*}

\subsection{Implementation details}

For the first task—dataset scoring using structural disparity on node level prediction—we employed a two-layer graph convolutional network (GCN) \cite{gcn} with a hidden dimension of 16. Additionally, we used a two-layer GraphSAGE \cite{sage}, also with a hidden dimension of 16, and a two-layer graph attention network (GAT) \cite{gat} with a hidden dimension of 16 and 8 attention heads. For clustering, we used the same GCN model with different parameters and applied KMeans \cite{kmeans} with 5 clusters. This task was evaluated across four different random seeds, and we report the average results.

For the second task—dataset scoring using relevance, diversity, and structural disparity—we utilized a three-layer GCN, where each hidden layer has a dimension of 64. We applied global mean pooling to aggregate node features into graph embeddings. The final layer is a multi-layer perceptron (MLP) with a dropout rate of \( p = 0.5 \). The network was trained end-to-end using the Adam optimizer \cite{adam}, with early stopping implemented to halt training if the validation loss did not improve for 25 consecutive epochs. The initial learning rate was set to \(10^{-2}\), and training was capped at 1000 epochs. A batch size of 32 was used for all datasets. Cross-entropy loss served as the loss function, and each dataset was evaluated using four random seeds. 
Experiments were conducted on a Windows machine equipped with an AMD Ryzen\texttrademark\ 7 4800HS processor (8-core/16-thread, 12MB Cache, 4.2 GHz max boost), an NVIDIA\textregistered\ GeForce RTX\texttrademark\ 2060 with Max-Q Design (6GB GDDR6), and 64GB of RAM.

\begin{figure*}[t!]
\centering
\includegraphics[width=1\textwidth]{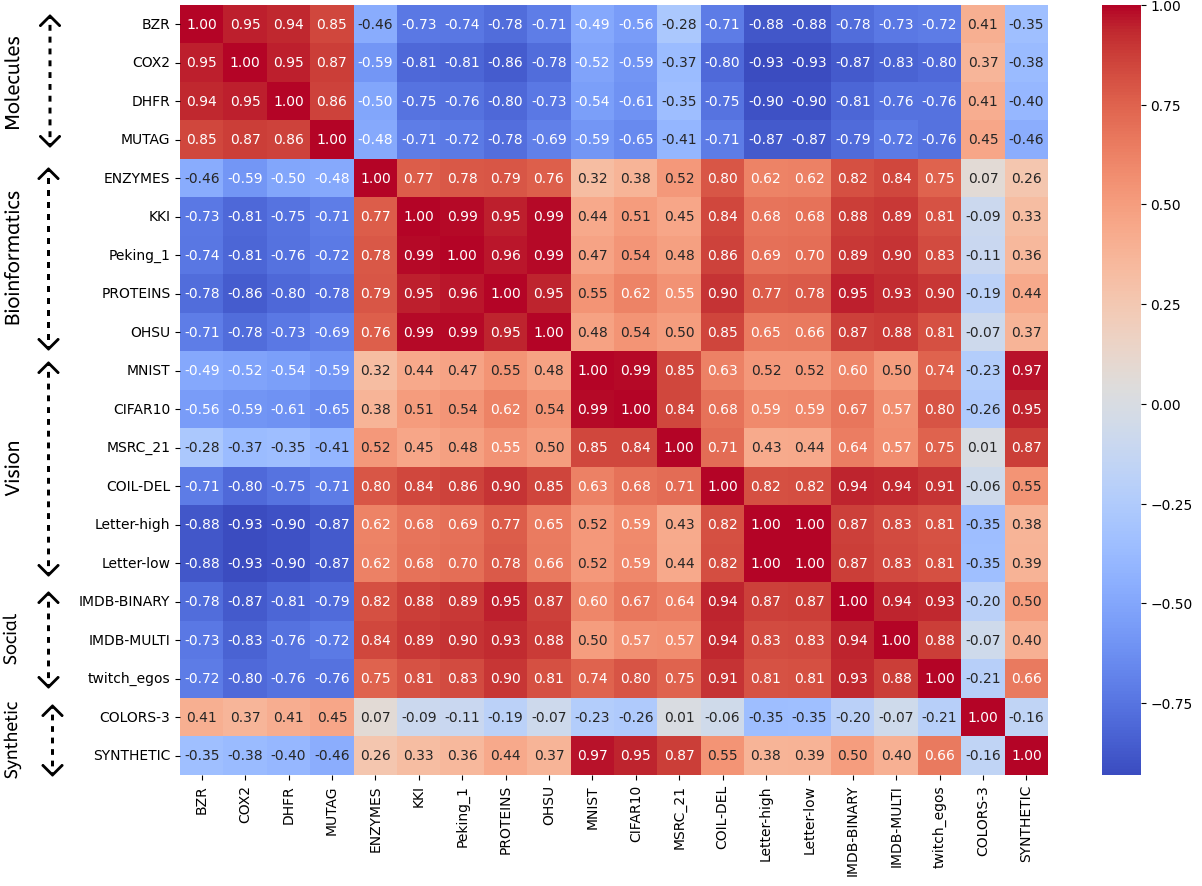}
\caption{\label{fig:corr} Pairwise score of datasets from five fields of molecules, bioinformatics, computer vision, social media, and synthetic datasets.}
\end{figure*}

\section{Additional experiments}

\begin{table*}[h]
\centering
\caption{\label{tab: ranks}Comparison of ranks with and without proxy across different datasets.}
\begin{tabular}{|c|c|c|c|c|c|c|c|c|c|}
\hline
 & \multicolumn{3}{c|}{PubMed} & \multicolumn{3}{c|}{Citeseer} & \multicolumn{3}{c|}{Cora} \\ \cline{2-10} 
   & R1 & R2 & R3 & R1 & R2 & R3 & R1 & R2 & R3 \\ \hline
Proxy    &         2      &         3      &         1      &        1       &        3       &           2    &            2   &        1       &     3          \\ \hline
No Proxy &      2         &          3     &        1       &       1        &       3        &         2      &            2   &      1         &      3         \\ \hline
\end{tabular}
\end{table*}

\begin{table*}[h]
\centering
\begin{tabular}{|c|c|c|c|c|c|c|c|c|c|c|c|c|c|c|c|}
\hline
 & \multicolumn{5}{c|}{PubMed} & \multicolumn{5}{c|}{Citeseer} & \multicolumn{5}{c|}{Cora} \\ \cline{2-16} 
               & R1 & R2 & R3 & R4 & R5
               & R1 & R2 & R3 & R4 & R5 
               & R1 & R2 & R3 & R4 & R5 \\ \hline
Proxy & 2           & 5           & 3           & 1           & 4           
               & 4           & 2           & 5           & 3           & 1           
               & 3           & 4           & 1           & 5           & 2           \\ \hline
No Proxy & 1        & 3           & 4           & 2           & 5           
                  & 5        & 1           & 2           & 4           & 3           
                  & 2        & 5           & 3           & 1           & 4           \\ \hline
\end{tabular}
\end{table*}

\subsection{(iii) Featural attributes in practice}

To evaluate the metrics diversity and relevance for capturing the featural attributes of the graphs, we conduct experiments on two datasets under various configurations. The results are summarized in Fig. \ref{fig:div}. The first dataset is the MNIST graph data from the GNNBenchmark dataset \cite{GNNBenchmark}. We create six distinct sets all with the same size sampled from the MNIST dataset. 
We examine a scenario where the buyers' graphs consist of only classes 0 to 4 from the MNIST dataset. There are five sellers, each offering graphs from MNIST but with different class ranges: classes 0 to 4 (matching the buyer), 1 to 5, 0 to 9, 3 to 9, and 5 to 9. It is evident that the diversity and relevance of the data should increase and decrease progressively from seller 1 to seller 5, a trend that our proposed estimates clearly capture. Notably, the seller providing data spanning all classes from 0 to 9 offers a diversity-relevance pair approximating the point (0.5, 0.5). This balanced position indicates that while the data includes a broad range of classes (increasing diversity), it still maintains moderate relevance to the buyer’s needs, as it covers the buyer's classes. To further validate our proposed estimates, we introduce a seller with a random dataset generated from a zero-mean, unit-variance Gaussian distribution, structured similarly to the other datasets. This seller exhibits the highest diversity and lowest relevance to the buyer, as expected because the random nature of the data significantly deviates from the buyer's classes. This setup is crucial as it highlights the effectiveness of our method in identifying data that, despite its high diversity, lacks practical utility for the buyer. Moreover, we observe that increasing the noise variance in the random dataset shifts the diversity-relevance pair closer to (1, 0). This shift is consistent with our theoretical expectations: as noise variance increases, the dataset's relevance to the buyer's graphs diminishes because the added noise distorts the data beyond what is useful for the buyer. Simultaneously, the diversity increases because the noise introduces more variability.

We further conduct this experiment using five distinct subsets from the FRANKENSTEIN dataset. The buyer's graphs exclusively consist of class 1 graphs. Each seller's dataset contains varying proportions of class 1 and class 2 graphs. Specifically, the first seller's dataset comprises 80\% class 1 and 20\% class 2, the second seller's dataset contains 60\% class 1 and 40\% class 2, the third seller's dataset includes 20\% class 1 and 80\% class 2, and the last seller's dataset is composed entirely of class 2 graphs. As expected, the last seller demonstrates the highest diversity and the lowest relevance to the buyer, given its 100\% composition of class 2 graphs, which contrasts completely with the buyer's class 1 graphs. As the percentage of class 2 graphs decreases in the sellers' datasets, their relevance to the buyer increases, while their diversity decreases, illustrating a trade-off between these two metrics.

The experiment effectively captures the trade-off between diversity and relevance in datasets offered by different sellers. As the composition of the datasets changes, the diversity and relevance metrics vary predictably, showcasing the robustness and sensitivity of the proposed metrics. This experiment validates the proposed diversity and relevance metrics by demonstrating their expected behavior across different configurations and datasets. It shows how these metrics can be used to assess the quality and suitability of datasets in scenarios where the buyer's and sellers' data distributions differ.

\subsection{ (iv) Structural context-awarenes}

Here we evaluate our proposed method to capture the distance between the structural attributes of graphs.
It is important to determine whether our proposed framework can effectively distinguish between graphs originating from different fields. To achieve this, we first create equal-sized sets sampled randomly from each of the datasets. We then embed each graph using the positional and structural embedders. Following this, we calculate the GWD between each pair of datasets. The datasets are categorized into five main groups: Molecules, Bioinformatics, Vision, Social, and Synthetic. The results of these calculations are summarized in Fig. \ref{fig:corr}.

Datasets in the Molecules category, such as BZR, COX2, DHFR, and MUTAG, show very high positive scores within their field and low scores with datasets from other fields. This indicates that graphs within the Molecules category have similar structural properties. Similarly, datasets from the Bioinformatics category, including KKI, Peking, PROTEINS, and OHSU, display high correlations within their group. However, they tend to have a high similarity with some of the social media datasets (IMDB-BINARY and IMDB-MULTI). The varied scores demonstrate that our framework can effectively differentiate between graphs from different fields. High scores within categories confirm the framework's ability to recognize similar graph structures, while low scores between categories highlight its capacity to distinguish disparate graph structures. Nonetheless, there are some limitations in distinguishing between the Vision datasets and the social media datasets, as indicated by their higher-than-expected scores. This can be attributed to the relatively low average number of nodes in some of the datasets within the social media and vision groups (IMDB-BINARY: 19.77, IMDB-MULTI: 13.00, Letter-high: 4.67, Letter-low: 4.68). Given that our embeddings, particularly LapPE, are sensitive to the number of nodes, this sensitivity could be a limiting factor affecting performance on datasets with a low number of nodes.

\subsection{ (v) Effects of using a proxy graph}  
To evaluate the impact of using a proxy graph like \(G_{\text{key}}\) on the performance of the BMP framework, it is essential to examine whether both approaches—employing the proxy graph to compute the optimal permutation versus directly finding the optimal permutation—select the sets of graphs for the buyer in the same order. This analysis is conducted using the same graph datasets as in the initial task of our experiment, which involved scoring datasets based on structural disparity in node level prediction. We begin by creating 4 and 6 subgraphs through node shuffling. We then assess the structural disparity between the first subgraph which we call baseline and the remaining subgraphs, ranking them based on proximity. For instance, if subgraph 2 has the highest structural similarity to the baseline, then subgraph 2 would be ranked as Rank 1 (R1) in Table \ref{tab: ranks}. Our objective is to compare the structural disparity calculated for each candidate set using the two approaches: with and without the proxy graph. In the proxy-based approach, we first determine the optimal permutation between the baseline and the proxy graph, as well as between each of the subgraphs and the proxy graph. Subsequently, we compute the GWD between them. However, in the without proxy approach, we compute the GWD directly between the baseline and each of the subgraphs without utilizing the proxy graph. The results are summarized in Table \ref{tab: ranks}. As observed, the use of proxy graphs consistently yields the same rankings as the approach without a proxy. This consistency supports our hypothesis that employing a proxy graph does not directly impact the performance of the BMP framework.

\section{Discussions} 
Here we discuss various aspects of the proposed approach, ranging from its properties to possible extensions.

\begin{enumerate}
    \item \textbf{Scalability}: While our method is effective for moderate-sized graphs, its scalability to extremely large graphs remains untested. The computational complexity of GWD, which is \(O(N^3)\), may pose challenges for very large datasets. Additionally, the matrix padding in computing \(\Phi\) can result in very sparse matrices, potentially leading to inefficiencies. Future research should explore techniques such as graph coarsening, parallel computing, and approximate algorithms to enhance scalability.

    \item \textbf{Feature incorporation}: Our current approach only incorporates either featural attributes related to nodes or edges. In scenarios where both nodes and edges have features, our method falls short. Further investigation is needed to extend our approach to heterogeneous graphs and hypergraphs, where both node and edge features can be simultaneously considered. This would make the method more versatile and applicable to a wider range of datasets.

    \item \textbf{Dynamic graphs}: Our current approach is designed for static graphs. Many real-world applications involve dynamic graphs that evolve over time, such as social networks, financial transaction networks, and communication networks. Extending our methodology to handle dynamic graphs would involve developing methods to track and adapt to changes in graph structure and features over time, significantly enhancing its applicability and robustness in real-time environments.

    \item \textbf{Privacy concerns}: While we emphasize privacy through the BPM framework, the approach still requires sharing some structural information, which could potentially lead to sensitive information leakage. Future work should explore incorporating more advanced privacy-preserving techniques, such as secure multi-party computation, differential privacy, and homomorphic encryption, into the BMP framework. These techniques can ensure that graph valuations can be performed without compromising sensitive information, thereby making the approach more secure.

    \item \textbf{Incorporating additional features}: Introducing more metrics beyond diversity, relevance, and structural disparity could enhance the data valuation process. This would provide a more comprehensive assessment of the data's value.
    
    \item \textbf{Incorporating utility functions}: In this paper, we did not introduce specific utility functions to maintain generalizability. Future work could explore incorporating these metrics into a utility function tailored to the specific context or task.
\end{enumerate}

\end{document}